\documentclass{article}
\usepackage{microtype}
\usepackage{booktabs}
\usepackage[utf8x]{inputenc}
\usepackage{times}
\usepackage{graphicx}
\usepackage{subfigure}
\usepackage{natbib}
\usepackage{algorithm}
\usepackage{algorithmic}
\usepackage{hyperref}

\usepackage[accepted]{icml2015}
\usepackage{multirow}
\usepackage{multicol}
\usepackage{color}
\usepackage{array}
\usepackage{amsmath}
\usepackage{amssymb}
\usepackage{standalone}
\newcommand{\sigm}{\mathrm{sigm}}

\newcommand{\te}{\!=\!}
\newcommand{\tp}{\!+\!}
\newcommand{\tm}{\!-\!}
\newcommand{\g}{\,|\,}

\newcommand{\bx}{{\bf x}}
\newcommand{\bh}{{\bf h}}
\newcommand{\bW}{{\bf W}}
\newcommand{\bM}{{\bf M}}
\newcommand{\bU}{{\bf U}}

\newcommand{\bV}{{\bf V}}
\newcommand{\bA}{{\bf A}}
\newcommand{\ba}{{\bf a}}
\newcommand{\bb}{{\bf b}}
\newcommand{\bc}{{\bf c}}

\newcommand{\bg}{{\bf g}}
\newcommand{\bm}{{\bf m}}
\newcommand{\bone}{{\bf 1}}

\icmltitlerunning{MADE: Masked Autoencoder for Distribution Estimation}

\begin{document}

\twocolumn[
\icmltitle{MADE: Masked Autoencoder for Distribution Estimation}

\icmlauthor{Mathieu Germain}{mathieu.germain2@usherbrooke.ca}
\icmladdress{Université de Sherbrooke, Canada}
\icmlauthor{Karol Gregor}{karol.gregor@gmail.com}
\icmladdress{Google DeepMind}
\icmlauthor{Iain Murray}{i.murray@ed.ac.uk}
\icmladdress{University of Edinburgh, United Kingdom}
\icmlauthor{Hugo Larochelle}{hugo.larochelle@usherbrooke.ca}
\icmladdress{Université de Sherbrooke, Canada}

\icmlkeywords{deep learning, distribution estimation, autoregressive modelling, machine learning, autoencoder, unsupervised learning, generative model, ICML}

\vskip 0.3in
]

\begin{abstract}
There has been a lot of recent interest in designing neural network models to estimate a distribution from a set of examples.
We introduce a simple modification for autoencoder neural networks that yields powerful generative models.
Our method masks the autoencoder's parameters to respect autoregressive constraints: each input is reconstructed only from previous inputs in a given ordering.
Constrained this way, the autoencoder outputs can be interpreted as a set of conditional probabilities, and their product, the full joint probability.
We can also train a single network that can decompose the joint probability in multiple different orderings.
Our simple framework can be applied to multiple architectures, including deep ones. Vectorized implementations, such as on GPUs, are simple and fast.
Experiments demonstrate that this approach is competitive with state-of-the-art tractable distribution estimators.
At test time, the method is significantly faster and scales better than other autoregressive estimators.
\end{abstract}

\section{Introduction}

Distribution estimation is the task of estimating a joint distribution $p(\bx)$ from a set of examples $\{\bx^{(t)}\}_{t=1}^T$, which is by definition a general problem. Many tasks in machine learning can be formulated as learning only specific properties of a joint distribution. Thus a good distribution estimator can be used in many scenarios, including classification~\cite{SchmahT2008}, denoising or missing input imputation~\cite{PoonD2011,DinhL2014}, data (e.g.\ speech) synthesis~\cite{UriaB2015} and many others. The very nature of distribution estimation also makes it a particular challenge for machine learning. In essence, the curse of dimensionality has a distinct impact because, as the number of dimensions of the input space of $\bx$ grows, the volume of space in which the model must provide a good answer for $p(\bx)$ exponentially increases.

Fortunately, recent research has made substantial progress on this task. Specifically, learning algorithms for a variety of neural network models have been proposed~\cite{BengioY1999,Larochelle+Murray-2011,GregorK2011,UriaB2013,UriaB2014,KingmaD2014,RezendeD2014,BengioY2014,GregorK2014,GoodfellowI2014,DinhL2014}. These algorithms are showing great potential in scaling to high-dimensional distribution estimation problems.
In this work, we focus our attention on \emph{autoregressive} models (Section~\ref{sec:ar}). Computing $p(\bx)$ exactly for a test example $\bx$ is tractable with these models. However, the computational cost of this operation is still larger than typical neural network predictions for a $D$-dimensional input. For previous deep autoregressive models, evaluating $p(\bx)$ costs $O(D)$ times more than a simple neural network point predictor.

This paper's contribution is to describe and explore a simple way of adapting autoencoder neural networks that makes them competitive tractable distribution estimators that are faster than existing alternatives. We show how to mask the weighted connections of a standard autoencoder to convert it into a distribution estimator. The key is to use masks that are designed in such a way that the output is autoregressive for a given ordering of the inputs, i.e.\ that each input dimension is reconstructed solely from the dimensions preceding it in the ordering.
The resulting Masked Autoencoder Distribution Estimator (MADE) preserves the efficiency of a single pass through a regular autoencoder.
Implementation on a GPU is straightforward, making the method scalable.

The single hidden layer version of MADE corresponds to the previously proposed autoregressive neural network of~\citet{BengioY1999}. Here, we go further by exploring deep variants of the model. We also explore training MADE to work simultaneously with multiple orderings of the input observations and hidden layer connectivity structures.
We test these extensions across a range of binary datasets with hundreds of
dimensions, and compare its statistical performance and scaling to comparable
methods.

\section{Autoencoders}

A brief description of the basic autoencoder, on which this work builds upon, is required to clearly grasp what follows. In this paper, we assume that we are given a training set of examples $\{\bx^{(t)}\}_{t=1}^T$. We concentrate on the case of binary observations, where for every $D$-dimensional input $\bx$, each input dimension $x_d$ belongs in $\{0,1\}$. The motivation is to learn hidden representations of the inputs that reveal the statistical structure of the distribution that generated them.

An autoencoder attempts to learn a feed-forward, hidden representation $\bh(\bx)$
of its input $\bx$ such that, from it, we can obtain a reconstruction $\widehat{\bx}$ which is as close as possible to $\bx$. Specifically, we have
\begin{eqnarray}
    \bh(\bx) &= & \bg(\bb + \bW \bx)\\
    \widehat{\bx} &= & \sigm(\bc + \bV \bh(\bx)) \,,
\end{eqnarray}
where $\bW$ and $\bV$ are matrices, $\bb$ and $\bc$ are vectors,
$\bg$ is a nonlinear activation function and $\sigm(a)=1/(1+\exp(-a))$.
Thus, $\bW$ represents the connections
from the input to the hidden layer, and $\bV$ represents the connections
from the hidden to the output layer.

To train the autoencoder, we must first specify a training loss function. For
binary observations, a natural choice is the cross-entropy loss:
\begin{equation}
  \ell(\bx) = \sum_{d=1}^D - x_d \log \widehat{x}_d - (1\tm x_d) \log (1\tm\widehat{x}_d)~.\label{eqn:loss_autoencoder}
\end{equation}
By treating $\widehat{x}_d$ as the model's probability that $x_d$ is 1, the
cross-entropy can be understood as taking the form of a negative log-likelihood
function. Training the autoencoder corresponds to optimizing the parameters
$\{\bW,\bV,\bb,\bc\}$ to reduce the average loss on the training examples,
usually with (mini-batch) stochastic gradient descent.

One advantage of the autoencoder paradigm is its
flexibility. In particular, it is straightforward to obtain a deep
autoencoder by inserting more hidden layers between the input and
output layers. Its main disadvantage is that the representation
it learns can be trivial. For instance, if the hidden layer
is at least as large as the input, hidden units can each
learn to ``copy'' a single input dimension, so as to reconstruct
all inputs perfectly at the output layer. One obvious
consequence of this observation is that the loss function
of Equation~\ref{eqn:loss_autoencoder} isn't in fact
a proper log-likelihood function. Indeed, since perfect reconstruction could be achieved, the implied data `distribution'
$q(\bx)\te \prod_d \widehat{x}_d^{x_d}(1\tm \widehat{x}_d)^{1-x_d}$ could be learned to be $1$ for any $\bx$ and thus not be properly normalized ($\sum_\bx q(\bx)\!\ne\! 1$).

\section{Distribution Estimation as Autoregression}
\label{sec:ar}

An interesting question is what property we could impose on the autoencoder, such that its output can be used to obtain valid probabilities. Specifically, we'd like to be able to write $p(\bx)$ in
 such a way that it could be computed based on the
 output of a properly corrected autoencoder.

 First, we can use the fact that, for any distribution, the probability
 product rule implies that we can always
 decompose it into the product of its nested conditionals

\begin{equation}
  p(\bx) = \prod_{d=1}^D p(x_d\g\bx_{<d}),%
\end{equation}
where $\bx_{<d} = [x_1,\dots,x_{d-1}]^\top$.

By defining $p(x_d\te1\g\bx_{<d})=\hat{x}_d$, and thus
$p(x_d\te0\g\bx_{<d})=1\tm\hat{x}_d$, the loss of
Equation~\ref{eqn:loss_autoencoder} becomes a valid negative
log-likelihood:
\begin{equation}
\begin{split}
  -\log p(\bx) &= \sum_{d=1}^D -\log p(x_d\g \bx_{<d})\\
&= \sum_{d=1}^D - x_d \log p(x_d\te1\g \bx_{<d}) \\
&~~~~~~~~~~~- (1\tm x_d) \log p(x_d\te0\g \bx_{<d}) \\
&=  \ell(\bx)\,.\label{eqn:autoreg_loss}
 \end{split}
\end{equation}

This connection provides a way to define autoencoders that can be used for distribution estimation.
Each output $\widehat{x}_d\te p(x_d\g\bx_{<d})$ must be a function taking as input $\bx_{<d}$ only and outputting the probability of observing value $x_d$ at the $d^{\rm th}$ dimension.
In particular, the autoencoder forms a proper distribution if each output unit $\hat{x}_d$ only depends on the previous input units $\bx_{<d}$, and not the other units $\bx_{\geq d} = [x_d, \dots, x_D]^\top$.

We refer to this property as the \emph{autoregressive property},
because computing the negative log-likelihood \eqref{eqn:autoreg_loss} is equivalent to sequentially predicting (regressing) each dimension of input~$\bx$.

\section{Masked Autoencoders}

The question now is how to modify
the autoencoder so as to satisfy the autoregressive property. Since output $\hat{x}_d$ must depend only
on the preceding inputs $\bx_{<d}$, it means that there
must be no computational path between output unit $\hat{x}_d$
and any of the input units $x_d,\dots,x_D$. In other
words, for each of these paths, at least one connection
(in matrix $\bW$ or $\bV$) must be 0.

A convenient way of zeroing connections is to elementwise-multiply
each matrix by a binary mask matrix, whose entries that are set to 0
correspond to the connections we wish to remove. For a single hidden layer
autoencoder, we write
\begin{eqnarray}
\bh(\bx) &=& \bg(\bb + (\bW \odot \bM^{\bW} )\bx)\\
\hat{\bx} &=&\sigm(\bc + (\bV \odot \bM^{\bV} ) \bh(\bx))
\end{eqnarray}
where $\bM^{\bW}$ and $\bM^{\bV}$ are the masks for $\bW$ and $\bV$ respectively.
It is thus left to the masks $\bM^{\bW}$ and $\bM^{\bV}$ to satisfy the
autoregressive property.

To impose the autoregressive property we first assign each unit in the
hidden layer an integer $m$ between $1$ and $D\tm1$ inclusively. The
$k^{\rm th}$ hidden unit's number $m(k)$ gives the maximum number of input
units to which it can be connected.
We disallow $m(k)\te D$ since this hidden unit would depend on all inputs
and could not be used in modelling any of the conditionals
$p(x_d\g\bx_{<d})$. Similarly, we exclude $m(k)\te0$, as it would
create constant hidden units.

The constraints on the maximum number of inputs to each hidden unit are encoded
in the matrix masking the connections between the input and hidden units:
\begin{equation}
   M^{\bW}_{k,d} = 1_{m(k)\ge d} = \left\{ \begin{array}{cl} 1 &\mbox{ if } m(k)\ge d \\ 0 &\mbox{ otherwise, } \end{array} \right. \label{eqn:hid_masks}
\end{equation}
for $d\!\in\!\{1,\dots,D\}$ and $k\!\in\!\{1,\dots,K\}$.
Overall, we need to encode the constraint
that the $d^{\rm th}$ output unit is only connected to
$\bx_{< d}$ (and thus not to $\bx_{\geq d}$). Therefore the output weights
can only connect the $d^{\rm th}$ output
to hidden units with $m(k)\!<\!d$, i.e.\ units
that are connected to at most $d\tm1$ input units.
These constraints are encoded in the output mask matrix:
\begin{equation}
   M^{\bV}_{d,k} = 1_{d>m(k)} = \left\{ \begin{array}{cl} 1 &\mbox{ if } d> m(k) \\ 0 &\mbox{ otherwise, } \end{array} \right.
\end{equation}
again for $d\in\{1,\dots,D\}$ and $k\in\{1,\dots,K\}$.
Notice that, from this rule, no hidden units will be connected to the
first output unit $\hat{x}_1$, as desired.

From these mask constructions, we can easily demonstrate that the corresponding
masked autoencoder satisfies the autoregressive property. First, we note that, since
the masks $\bM^{\bV}$ and $\bM^{\bW}$ represent the network's connectivity,
their matrix product $\bM^{\bV,\bW} = \bM^{\bV}\bM^{\bW}$ represents
the connectivity between the input and the output layer. Specifically,
$M^{\bV,\bW}_{d',d}$ is the number of network paths between
output unit $\hat{x}_{d'}$ and
input unit $x_d$.
Thus, to demonstrate the autoregressive property, we need to show that
$\bM^{\bV,\bW}$ is strictly lower diagonal, i.e.\ $M^{\bV,\bW}_{d',d}$
is 0 if $d'\!\leq\! d$. By definition of the matrix product, we have:
\begin{equation}
  M^{\bV,\bW}_{d',d} = \sum_{k=1}^K M^{\bV}_{d',k} M^{\bW}_{k,d} =  \sum_{k=1}^K 1_{d'>m(k)} 1_{m(k)\ge d}.
\end{equation}
If $d'\!\leq\! d$, then there are no values for $m(k)$ such that it is both strictly less than
$d'$ and greater or equal to $d$. Thus $M^{\bV,\bW}_{d',d}$ is indeed 0.

Constructing the masks $\bM^{\bV}$ and $\bM^{\bW}$ only requires an
assignment of the $m(k)$ values to each hidden unit. One could imagine trying to
assign an (approximately) equal number of units to each legal value of $m(k)$.
In our experiments, we instead set $m(k)$ by sampling from a uniform discrete
distribution defined
on integers from $1$ to $D\tm1$, independently for each of the $K$ hidden units.

Previous work on autoregressive neural networks have also found it advantageous to
use direct connections between the input and output layers~\cite{BengioY1999}.
In this context, the reconstruction becomes:
\begin{equation}
    \hat{\bx} = \sigm(\bc + (\bV \odot \bM^{\bV} )\bh(\bx) + (\bA \odot \bM^{\bA} ) \bx )\,,
\end{equation}
where $\bA$ is the parameter connection matrix and $\bM^{\bA}$ is its mask matrix.
To satisfy the autoregressive property, $\bM^{\bA}$ simply needs to be a strictly lower diagonal
matrix, filled otherwise with ones.
We used such direct connections in our experiments as well.

\subsection{Deep MADE}

One advantage of the masked autoencoder framework described in the previous section
is that it naturally generalizes to deep architectures.
Indeed, as we'll see, by assigning a maximum number of connected inputs to
all units across the deep network, masks can be similarly constructed
so as to satisfy the autoregressive property.

\begin{figure}[t]
    \begin{center}
    \begin{minipage}{0.49\textwidth}
    \includegraphics[width=0.99\linewidth]{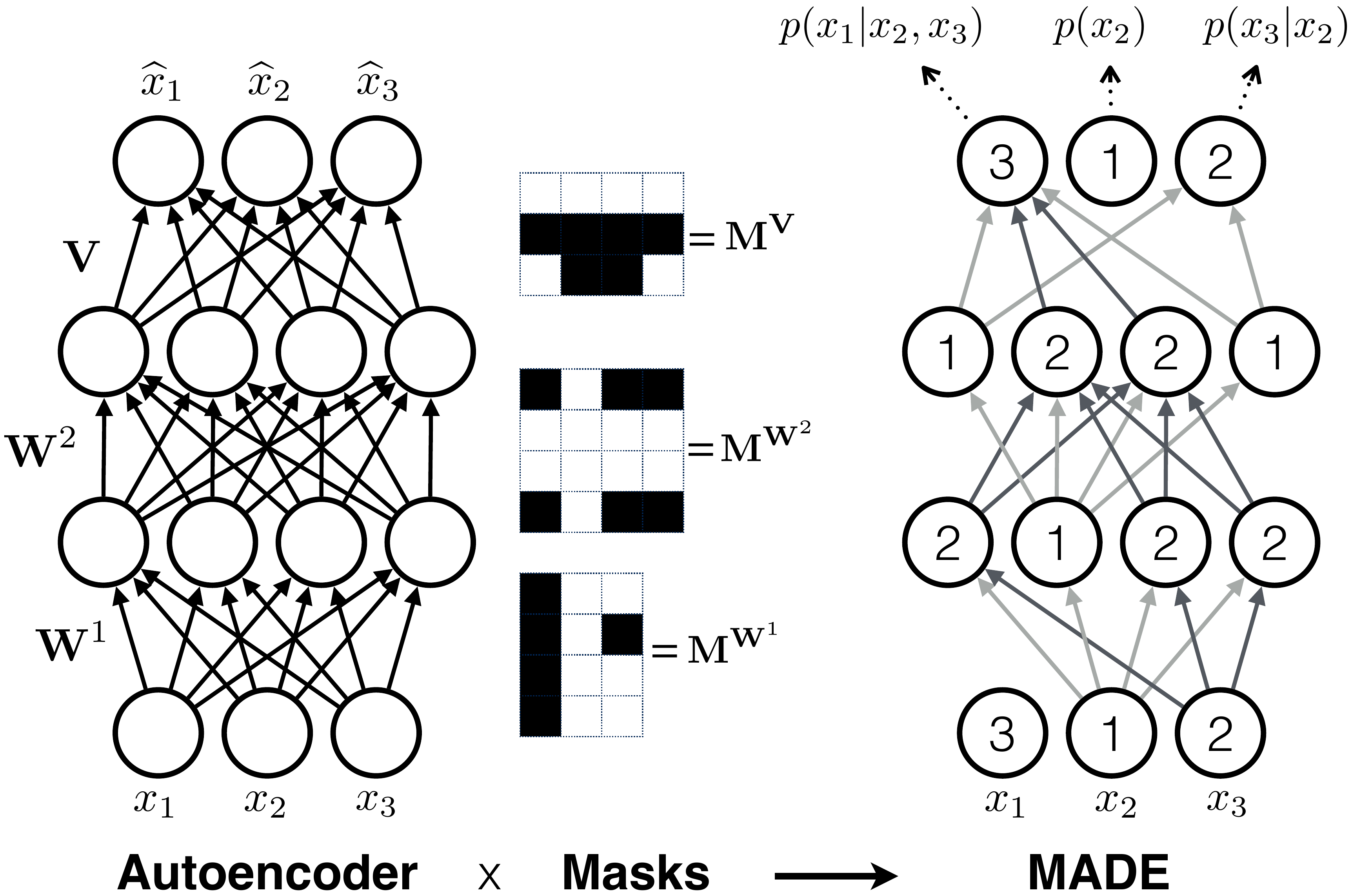}
    \end{minipage}
    \end{center}
    \caption{\textbf{Left: Conventional three hidden layer autoencoder}. Input in the bottom is passed through fully connected layers and point-wise nonlinearities. In the final top layer, a reconstruction specified as a probability distribution over inputs is produced. As this distribution depends on the input itself, a standard autoencoder cannot predict or sample new data. \textbf{Right: MADE}\@. The network has the same structure as the autoencoder, but a set of connections is removed such that each input unit is only predicted from the previous ones, using multiplicative binary masks (\smash{$\bM^{\bW^1}$, $\bM^{\bW^2}$, $\bM^{\bV}$}). In this example, the ordering of the input is changed from 1,2,3 to 3,1,2. This change is explained in section~\ref{subsection:order_agnostic}, but is not necessary for understanding the basic principle. The numbers in the hidden units indicate the maximum number of inputs on which the $k^{\rm th}$ unit of layer $l$ depends. The masks are constructed based on these numbers (see Equations~\ref{eqn:hid_masks_deep}~and~\ref{eqn:out_masks_deep}). These masks ensure that MADE satisfies the autoregressive property, allowing it to form a probabilistic model, in this example $p(\bx) = p(x_2)\, p(x_3|x_2)\, p(x_1|x_2,x_3)$. Connections in light gray correspond to paths that depend only on 1 input, while the dark gray connections depend on 2 inputs.}
    \label{fig:made}
\end{figure}

For networks with $L\!>\!1$ hidden layers, we use superscripts to
index the layers.
The first hidden layer matrix (previously $\bW$) will be denoted
$\bW^{1}$, the second hidden layer matrix will be $\bW^2$, and so on.
The number of hidden units (previously $K$) in each hidden layer will
be similarly indexed as $K^l$, where $l$ is the hidden layer index.
We will also generalize the notation for the maximum number of connected
inputs of the $k^{\rm th}$ unit in the $l^{\rm th}$ layer to $m^l(k)$.

We've already discussed how to define the first layer's mask
matrix such that it ensures that its $k^{\rm th}$ unit is connected
to at most $m(k)$ (now $m^1(k)$) inputs. To impose the
same property on the second hidden layer, we must simply make sure
that each unit $k'$ is only connected to first layer units
connected to at most $m^2(k')$ inputs, i.e.\ the first layer units such that
$m^1(k)\leq m^2(k')$.

One can generalize this rule to any layer $l$, as follows:
\begin{equation}
   M^{\bW^{l}}_{k',k} = 1_{m^l(k')\ge m^{l-1}(k)} = \left\{ \begin{array}{cl} 1 &\mbox{ if } m^l(k')\ge m^{l-1}(k) \\ 0 &\mbox{ otherwise. } \end{array} \right. \label{eqn:hid_masks_deep}
\end{equation}
Also, taking $l\te0$ to mean the input layer and defining $m^0(d) \te d$ (which is intuitive,
since the $d^{\rm th}$ input unit indeed takes its values from the $d$ first inputs),
this definition also applies for the first hidden layer weights.
As for the output mask, we simply need to adapt its definition by using the
connectivity constraints of the last hidden layer $m^{L}(k)$ instead of the first:
\begin{equation}
   M^{\bV}_{d,k} = 1_{d>m^{L}(k)} = \left\{ \begin{array}{cl} 1 &\mbox{ if } d> m^{L}(k) \\ 0 &\mbox{ otherwise. } \end{array} \right. \label{eqn:out_masks_deep} \end{equation}
Like for the single hidden layer case, the values for $m^{l}(k)$ for each hidden layer $l\in\{1,\dots,L\}$
are sampled uniformly.
To avoid unconnected units, the
value for $m^{l}(k)$ is sampled to be greater than
or equal to the minimum connectivity at the previous layer, i.e.\ $\min_{k'} m^{l-1}(k')$.

\subsection{Order-agnostic training}
\label{subsection:order_agnostic}
So far, we've assumed that the conditionals modelled by
MADE were consistent with the natural ordering of the
dimensions of $\bx$. However, we might be interested
in modelling the conditionals associated with an
arbitrary ordering of the input's dimensions.

Specifically, \citet{UriaB2014} have shown
that training an autoregressive model on {\it all}
orderings can be beneficial. We refer to this approach
as order-agnostic training. It can be achieved
by sampling an ordering before each stochastic/minibatch gradient
update of the model. There are two advantages of this approach.
Firstly, missing values in partially observed input vectors can be
imputed efficiently:
we invoke an ordering where observed dimensions are
all before unobserved ones, making inference straightforward.
Secondly, an ensemble of autoregressive models can
be constructed on the fly, by exploiting the fact that the
conditionals for two different orderings are not guaranteed
to be exactly consistent (and thus technically correspond to
slightly different models). An ensemble is then easily obtained
by sampling a set of orderings, computing the probability of $\bx$ under
each ordering and averaging.

Conveniently, in MADE, the ordering is simply represented by
the vector $\bm^0 = [m^{0}(1),\dots,m^{0}(D)]$. Specifically, $m^{0}(d)$
corresponds to the position of the original $d^{\rm th}$ dimension of
$\bx$ in the product of conditionals. Thus, a random ordering can
be obtained by randomly permuting the ordered vector $[1,\dots,D]$.
From these values of each $\bm^0$, the first hidden layer
mask matrix can then be created.
During order-agnostic training, randomly permuting
the last value of $\bm^0$ again is sufficient to obtain a new
random ordering.

\subsection{Connectivity-agnostic training}

One advantage of order-agnostic training is that it effectively
allows us to train as many models as there are orderings, using
a common set of parameters. This can be exploited by creating
ensembles of models at test time.

\begin{algorithm}[ht!]
   \caption{Computation of $p(\bx)$ and learning gradients for MADE with order and connectivity sampling. $D$ is the size of the input, $L$ the number of hidden layers and $K$ the number of hidden units.}
   \label{alg:made-fbprop}
\begin{algorithmic}
\smallskip
   \STATE {\bfseries Input:} training observation vector $\bx$
   \STATE {\bfseries Output:} $p(\bx)$ and gradients of $-\log p(\bx)$ on parameters
   \STATE
   \STATE \# Sampling $\bm^l$ vectors
   \STATE $\bm^0 \leftarrow {\rm shuffle}([1,\dots,D])$
   \FOR{$l$ from 1 to $L$}
   \FOR{$k$ from 1 to $K^l$}
   \STATE $m^l(k) \leftarrow {\rm Uniform}([\min_{k'} m^{l-1}(k'),\dots,D\!-\!1])$
   \ENDFOR
   \ENDFOR
   \STATE
   \STATE \# Constructing masks for each layer
   \FOR{$l$ from 1 to $L$}
   \STATE $\bM^{\bW^l} \leftarrow \bone_{\bm^{l}\ge \bm^{l-1}}$
   \ENDFOR
   \STATE $\bM^{\bV} \leftarrow \bone_{\bm^0>\bm^{L}}$
   \STATE
   \STATE \# Computing $p(\bx)$
   \STATE $\bh^0(\bx) \leftarrow \bx$
   \FOR{$l$ from 1 to $L$}
   \STATE $\bh^l(\bx) \leftarrow \bg(\bb^l + (\bW^l \odot \bM^{\bW^l} )\bh^{l-1}(\bx))$
   \ENDFOR
   \STATE $\widehat{\bx} \leftarrow \sigm(\bc + (\bV \odot \bM^{\bV} )\bh^{L}(\bx))$
   \STATE $p(\bx) \leftarrow \exp\left(\sum_{d=1}^D x_d \log \widehat{x}_d +(1\!-\!x_d) \log (1\!-\!\widehat{x}_d)\right)$
   \STATE
   \STATE \# Computing gradients of $\,-\log p(\bx)$
   \STATE ${\rm tmp} \leftarrow \widehat{\bx}-\bx$
   \STATE $\delta \bc \leftarrow {\rm tmp}$
   \STATE $\delta \bV \leftarrow \left({\rm tmp}~ \bh^L(\bx)^\top\right)\odot \bM^{\bV}$
   \STATE ${\rm tmp} \leftarrow ({\rm tmp}^\top (\bV \odot \bM^{\bV}))^\top$
   \FOR{$l$ from $L$ to 1}
   \STATE ${\rm tmp} \leftarrow {\rm tmp} \odot \bg'(\bb^l + (\bW^l \odot \bM^{\bW^l} )\bh^{l-1}(\bx))$
   \STATE $\delta \bb^l \leftarrow {\rm tmp}$
   \STATE $\delta \bW^l \leftarrow \left({\rm tmp}~ \bh^{l-1}(\bx)^\top\right)\odot \bM^{\bW^l}$
   \STATE ${\rm tmp} \leftarrow ({\rm tmp}^\top (\bW^l \odot \bM^{\bW^l}))^\top$
   \ENDFOR
   \STATE {\bf return} $p(\bx), \delta \bb^1,\dots,\delta \bb^L, \delta \bW^1, \dots, \delta \bW^L, \delta \bc, \delta \bV$
\end{algorithmic}
\end{algorithm}

In MADE, in addition to choosing an ordering, we also have to
choose each hidden unit's connectivity constraint $m^l(k)$.
Thus, we could imaging training MADE to also be agnostic of the
connectivity pattern generated by these constraints. To
achieve this, instead of sampling the values of $m^l(k)$ for all
units and layers once and for all before training, we actually
resample them for each training example or minibatch.
This is still practical, since the operation of creating the masks
is easy to parallelize. Denoting $\bm^l = [m^l(1),\dots,m^l(K^l)]$, and
assuming an element-wise and parallel implementation of the operation $1_{a\ge b}$ for vectors,
such that ${\bone_{\ba\ge \bb}}$ is a matrix whose $i,j$ element is $1_{a_i\ge b_j}$, then the hidden layer
masks are simply $\bM^{\bW^{l}} = \bone_{\bm^{l}\ge \bm^{l-1}}$.

By resampling the connectivity of hidden units for every update,
each hidden unit will have a constantly changing number of incoming
inputs during training. However, the absence of a connection
is indistinguishable from an instantiated connection to a zero-valued unit,
which could confuse the neural network during training. In a similar situation, \citet{UriaB2014}
informed each hidden unit which units were providing input with binary indicator
variables, connected with additional learnable weights.
We considered applying a similar strategy,
using companion weight matrices $\bU^l$, that are also masked
by $\bM^{\bW^l}$ but connected to a constant
one-valued vector:
\begin{equation}
\bh^l(\bx) = \bg(\bb^l + (\bW^l \odot \bM^{\bW^l} )\bh^{l-1}(\bx) + (\bU^l \odot \bM^{\bW^l} )\bone)
\end{equation}
An analogous parametrization of the output layer was also employed. These
connectivity conditioning weights were only sometimes useful. In
our experiments, we treated the choice of using them as a hyperparameter.

Moreover, we've found in our experiments that sampling masks for every
example could sometimes over-regularize MADE and provoke underfitting.
To fix this issue, we also considered sampling from only a finite list of masks.
During training, MADE cycles through this list, using one for every update.
At test time, we then average probabilities obtained for all masks in the list.

Algorithm~\ref{alg:made-fbprop} details how $p(\bx)$ is computed
by MADE, as well as how to obtain the gradient of $\ell(\bx)$ for
stochastic gradient descent training. For simplicity, the pseudocode assumes
order-agnostic and connectivity-agnostic training, doesn't
assume the use of conditioning weight matrices
or of direct input/output connections.
Figure~\ref{fig:made} also illustrates an example of such a two-layer MADE network, along with its $m^l(k)$ values and its masks.

\section{Related Work}

There has been a lot of recent work on exploring the use of
feed-forward, autoencoder-like neural networks as probabilistic
generative models. Part of the motivation behind this research is to
test the common assumption that the use of models with probabilistic
latent variables and intractable partition functions (such as the
restricted Boltzmann machine~\cite{SalakhutdinovR2008}), is a
necessary evil in designing powerful generative models for
high-dimensional data.

The work on the neural autoregressive distribution estimator or NADE~\cite{Larochelle+Murray-2011} has
illustrated that feed-forward architectures can in fact be used to form
state-of-the-art and even tractable distribution estimators.

Recently, a deep extension of NADE was proposed, improving even further the
state-of-the-art in distribution estimation~\cite{UriaB2014}. This work
introduced a randomized training procedure, which (like MADE) has nearly the
same cost per iteration as a standard autoencoder. Unfortunately,
deep NADE models still require $D$
feed-forward passes through the network to evaluate the probability $p(\bx)$
of a $D$-dimensional test vector. The computation of the first hidden
layer's activations can be shared across these passes, although is slower in practice
than evaluating a single pass in a standard autoencoder. In deep
networks with $K$ hidden units per layer, it costs $O(DK^2)$ to evaluate a
test vector.

Deep AutoRegressive Networks \citep[DARN,][]{GregorK2014}, also provide
probabilistic models with roughly the same training costs as standard
autoencoders. DARN's latent representation consist of binary, stochastic hidden units. While simulating from these models is fast, evaluation of exact
test probabilities requires summing over all configurations of the latent representation, which is exponential in computation. Monte Carlo
approximation is thus recommended.

The main advantage of MADE is that evaluating probabilities retains the
efficiency of autoencoders, with minor additional cost for simple masking
operations. Table~\ref{tab:model_complexity} lists the
computational complexity for exact computation of probabilities for various
models. DARN and RBMs are exponential in dimensionality of the hiddens or data,
whereas NADE and MADE are polynomial. MADE only requires one pass through the
autoencoder rather than the $D$ passes required by NADE\@. In practice, we also observe that the single-layer MADE is an order of magnitude faster than a one-layer NADE, for the same hidden layer size, despite NADE
sharing computation to get the same asymptotic scaling. NADE's computations
cannot be vectorized as efficiently.
The deep versions of MADE also have better scaling than NADE at test time.
The training costs for MADE, DARN, and deep NADE will all be similar.

\begin{table}
\small
\begin{center}
\caption{Complexity of the different models in Table~\ref{tab:mnist_results}, to compute an exact test negative log-likelihood. $R$ is the number of orderings used, $D$ is the input size, and $K$ is the hidden layer size (assuming equally sized hidden layers).}
\smallskip
\begin{tabular}{lc}\toprule
    Model & ${O}\textrm{\tiny{NLL}}$ \\
\midrule
RBM 25 CD steps       & ${O}(\min(2^{D}K,\,D2^{K}))$ \\
DARN                  & \smash{${O}(2^K D)$} \\
NADE (fixed order)    & ${O}(DK)$  \\
EoNADE 1hl, $R$ ord.  & ${O}(RDK)$ \\
EoNADE 2hl, $R$ ord.  & ${O}(RDK^{2})$ \\
\midrule
MADE 1hl, 1 ord.      & ${O}(DK \tp D^2)$ \\
MADE 2hl, 1 ord.      & ${O}(DK \tp K^{2} \tp D^2)$ \\
MADE 1hl, $R$ ord.     & ${O}(R(DK \tp D^2))$ \\
MADE 2hl, $R$ ord.     & ${O}(R(DK \tp K^{2} \tp D^2))$ \\
\bottomrule
\label{tab:model_complexity}
\end{tabular}
\end{center}
\end{table}

Before the work on NADE, \citet{BengioY1999} proposed a neural
network architecture that corresponds to the special case of a
single hidden layer MADE model, without randomization of input
ordering and connectivity. A contribution
of our work is to go beyond this special case, exploring deep variants
and order/connectivity-agnostic training.

An interesting interpretation of the autoregressive mask sampling is
as a structured form of dropout regularization~\cite{SrivastavaN2014}. Specifically, it bears similarity
with the masking in dropconnect networks~\cite{WanL2013}. The exception is that
the masks generated here must guaranty the autoregressive property of the
autoencoder, while in \citet{WanL2013}, each element in the mask is generated
independently.

\section{Experiments}

To test the performance of our model we considered two different benchmarks: a suite of UCI binary datasets, and the binarized MNIST dataset.
The code to reproduce the experiments of this paper is available at https://github.com/mgermain/MADE/releases/tag/ICML2015.
The results reported here are the average negative log-likelihood on the test set of each respective dataset. All experiments were made using stochastic gradient descent (SGD) with mini-batches of size 100 and a lookahead of 30 for early stopping.

\subsection{UCI evaluation suite}

We use the binary UCI evaluation suite that was first put together in \citet{Larochelle+Murray-2011}. It's a collection of 7 relatively small datasets from the University of California, Irvine machine learning repository and the OCR-letters dataset from the Stanford AI Lab. Table~\ref{tab:uci_datasets_info} gives an overview of the scale of those datasets and the way they were split.

\begin{table}
\small
\begin{center}
\caption{Number of input dimensions and numbers of examples in the train, validation, and test splits.}
\smallskip
\begin{tabular}{lrrrr}\toprule
Name & \# Inputs & Train & Valid\rlap{.} & Test \\
\midrule
Adult       & 123 &  5000 &  1414 &  26147 \\
Connect4    & 126 & 16000 &  4000 &  47557 \\
DNA         & 180 &  1400 &   600 &   1186 \\
Mushrooms   & 112 &  2000 &   500 &   5624 \\
NIPS-0-12   & 500 &   400 &   100 &   1240 \\
OCR-letters & 128 & 32152 & 10000 &  10000 \\
RCV1        & 150 & 40000 & 10000 & 150000 \\
Web         & 300 & 14000 &  3188 &  32561 \\
\bottomrule
\label{tab:uci_datasets_info}
\end{tabular}
\end{center}
\end{table}

The experiments were run with networks of 500 units per hidden layer, using the adadelta learning update~\cite{Zeiler2012} with a decay of 0.95. The other hyperparameters were varied as Table~\ref{tab:hyperparams} indicates.
We note as \textit{\# of masks} the number of different masks through which MADE cycles during training. In the no limit case, masks are sampled on the fly and never explicitly reused unless re-sampled by chance. In this situation, at validation and test time, 300 and 1000 sampled masks are used for averaging probabilities.

\begin{table}
\small
\begin{center}
\caption{UCI Grid Search}
\smallskip
\begin{tabular}{ll}\toprule
Hyperparameter & Values tried \\
\midrule
\# Hidden Layer             & 1, 2 \\
Activation function         & ReLU, Softplus \\
Adadelta epsilon            & \smash{\hbox{$\text{10}^{-\text{5}}$, $\text{10}^{-\text{7}}$, $\text{10}^{-\text{9}}$}} \\
Conditioning Weights        & True, False\\
\# of orderings             & 1, 8, 16, 32, No Limit \\
\bottomrule
\label{tab:hyperparams}
\end{tabular}
\end{center}
\end{table}

The results are reported in Table~\ref{tab:results}. We see that MADE is among the best performing models on half of the datasets and is competitive otherwise. To reduce clutter, we have not reported standard deviations, which were fairly small and consistent across models. However, for completeness we report standard deviations in a separate table in the supplementary materials.

\begin{table*}[t]
\small
\begin{center}
\caption{Negative log-likelihood test results of different models on multiple datasets. The best result as well as any other result with an overlapping confidence interval is shown in bold. Note that since the variance of DARN was not available, we considered it to be zero.}
\smallskip
\begin{tabular}{lcccccccc}\toprule
Model       & Adult & Connect4 & DNA & Mushrooms & NIPS-0-12 & OCR-letters & RCV1 & Web \\
\midrule
MoBernoullis         & 20.44      & 23.41      & 98.19      & 14.46      & 290.02      & 40.56          & 47.59               & 30.16 \\
RBM                  & 16.26      & 22.66      & 96.74      & 15.15      & 277.37      & 43.05          & 48.88               & 29.38 \\
FVSBN                & \bf{13.17} & 12.39      & 83.64      & 10.27      & 276.88      & 39.30          & 49.84               & 29.35 \\
NADE (fixed order)   & \bf{13.19} & 11.99      & 84.81      &  9.81      & \bf{273.08} & \bf{27.22}     & 46.66               & 28.39 \\
EoNADE 1hl (16 ord.) & \bf{13.19} & 12.58      & 82.31      &  9.69      & \bf{272.39} & \bf{27.32}     & \bf{46.12}          & \bf{27.87} \\
DARN                 & 13.19      & 11.91      & 81.04      &  \bf{9.55} & 274.68      & $\approx$28.17 & $\approx$\bf{46.10} & $\approx$28.83 \\
\midrule
MADE                 & \bf{13.12} & \bf{11.90} & 83.63      &  9.68      & 280.25      & 28.34          & 47.10               & 28.53 \\
MADE mask sampling         & \bf{13.13} & \bf{11.90} & \bf{79.66} &  9.69      & 277.28      & 30.04          & 46.74               & \bf{28.25} \\
\bottomrule
\label{tab:results}
\end{tabular}
\end{center}
\end{table*}

An analysis of the hyperparameters selected for each dataset reveals no clear winner. However, we do see from Table~\ref{tab:results} that when the mask sampling helps, it helps quite a bit and when it does not, the impact is negligible on all but OCR-letters. Another interesting note is that the conditioning weights had almost no influence except on NIPS-0-12 where it helped.

\subsection{Binarized MNIST evaluation}

The version of MNIST we used is the one binarized by \citet{SalakhutdinovR2008}. MNIST is a set of 70,000 hand written digits of 28$\times$28 pixels. We use the same split as in \citet{Larochelle+Murray-2011}, consisting of 50,000 for the training set, 10,000 for the validation set and 10,000 for the test set.

Experiments were run using the adagrad learning update~\cite{Duchi:EECS-2010-24}, with an epsilon of $10^{-6}$. Since MADE is much more efficient than NADE, we considered varying the hidden layer size from 500 to 8000 units. Seeing that increasing the number of units tended to always help, we used 8000.
Even with such a large hidden layer, our GPU implementation of MADE was quite efficient. Using a single mask, one training epoch requires about 14 and 44 seconds, for one hidden layer and two hidden layer MADE respectively. Using 32 sampled masks, training time increases to 33 and 100 respectively. These timings are all less than our GPU implementation of the 500 hidden units NADE model, which requires about 130 seconds per epoch. These timings were obtained on a K20 NVIDIA GPU.

Building on what we learned on the UCI experiments, we set the activation function to be ReLU and the conditioning weights were not used. The hyperparameters that were varied are in Table~\ref{tab:hyperparams_mnist}.

\begin{table}
\small
\begin{center}
\caption{Binarized MNIST Grid Search}
\smallskip
\begin{tabular}{ll}\toprule
Hyperparameter & Values tried \\
\midrule
\# Hidden Layer             & 1, 2 \\
Learning Rate               & 0.1, 0.05, 0.01, 0.005 \\
\# of masks              & 1, 2, 4, 8, 16, 32, 64 \\
\bottomrule
\label{tab:hyperparams_mnist}
\end{tabular}
\end{center}
\end{table}

The results are reported in Table~\ref{tab:mnist_results}, alongside other results taken from the literature. Again, despite its tractability, MADE is competitive with other models. Of note is the fact that the best MADE model outperforms the single layer NADE network, which was otherwise the best model among those requiring only a single feed-forward pass to compute log probabilities.

\begin{figure}
    \vspace*{-0.4cm}
    \begin{center}
        \includegraphics[scale=0.45]{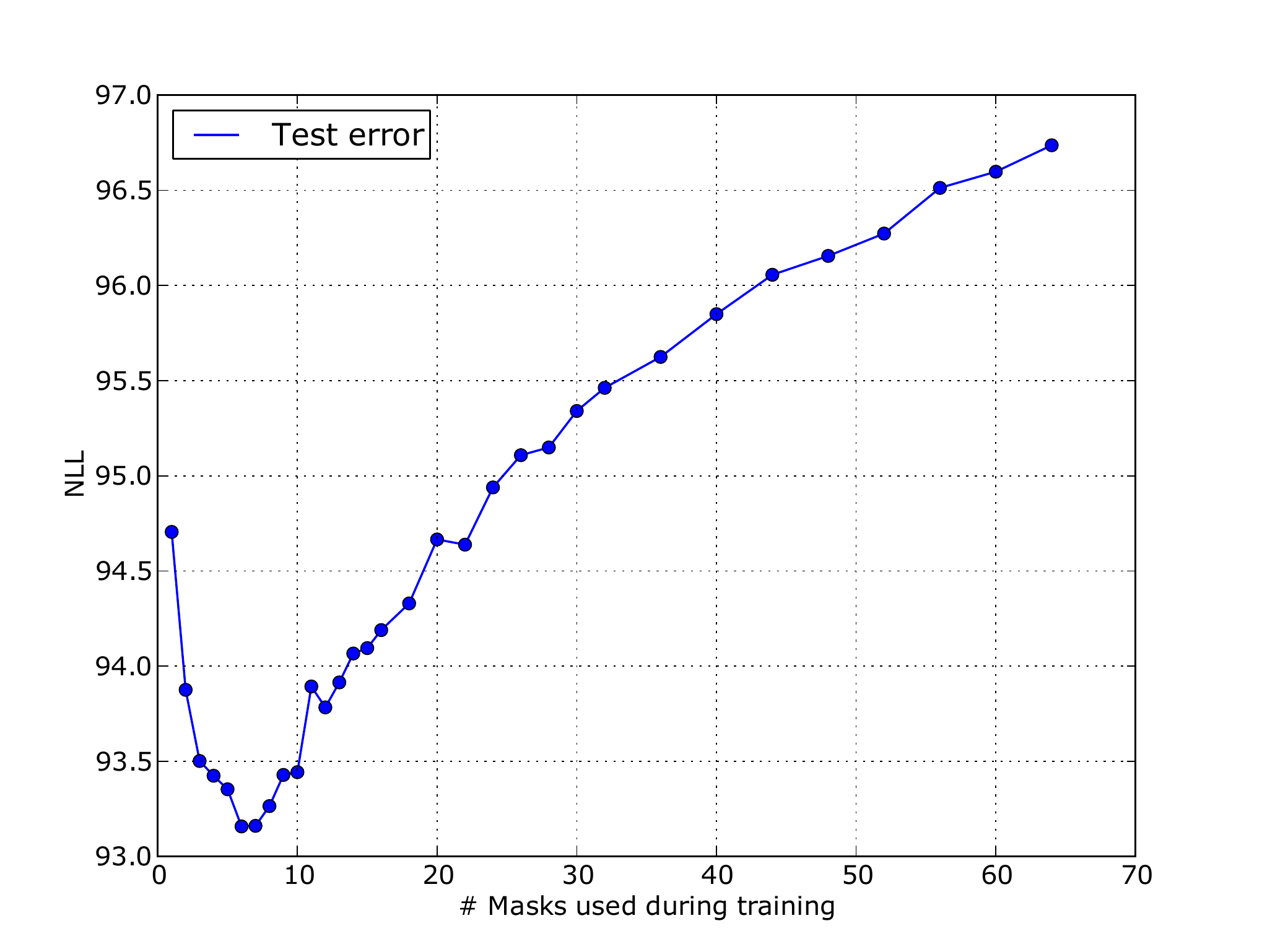}
        \caption{Impact of the number of masks used with a single hidden layer, 500 hidden units network, on binarized MNIST.}
        \label{fig:number_of_masks_plot}
    \end{center}
\end{figure}

In these experiments, we clearly observed the over-regularization phenomenon from using too many masks. When more than four orderings were used, the deeper variant of MADE always yielded better results. For the two layer model, adding masks during training helped up to 64, at which point the negative log-likelihood started to increase. We observed a similar pattern for the single layer model, but in this case the dip was around 8 masks. Figure~\ref{fig:number_of_masks_plot} illustrates this behaviour more precisely for a single layer MADE with 500 hidden units, trained by only varying the number of masks used and the size of the mini-batches (83, 100, 128).

\begin{table}
\small
\vspace*{-0.5cm}
\begin{center}
\caption{Negative log-likelihood test results of different models on the binarized MNIST dataset.}
\smallskip
\begin{tabular}{lcc}\toprule
Model                     & $−\log p$ & \\
\midrule
RBM (500 h, 25 CD steps)  & $\approx$ 86.34 & \multirow{5}{*}{\rotatebox[origin=c]{90}{Intractable}} \\
DBM 2hl                   & $\approx$ 84.62 \\
DBN 2hl                   & $\approx$ 84.55 \\
DARN $n_h$=500            & $\approx$ 84.71 \\
DARN $n_h$=500, adaNoise  & $\approx$ 84.13 \\
\midrule
MoBernoullis K=10         & 168.95 & \multirow{9}{*}{\rotatebox[origin=c]{90}{Tractable}} \\
MoBernoullis K=500        & 137.64 \\
NADE 1hl (fixed order)    & 88.33 \\
EoNADE 1hl (128 orderings)& 87.71 \\
EoNADE 2hl (128 orderings)& 85.10 \\
\cmidrule{1-2}
MADE 1hl (1 mask)     & 88.40 \\
MADE 2hl (1 mask)     & 89.59 \\
MADE 1hl (32 masks)   & 88.04 \\
MADE 2hl (32 masks)   & 86.64 \\
\bottomrule
\label{tab:mnist_results}
\end{tabular}
\end{center}
\end{table}

\begin{figure*}
    \begin{center}
        \includegraphics[scale=0.91]{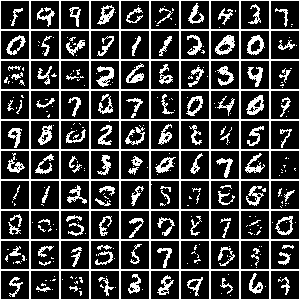}\quad
        \includegraphics[scale=0.91]{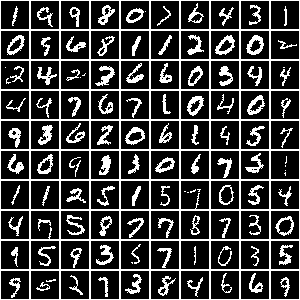}
        \caption{Left: Samples from a 2 hidden layer MADE\@. Right: Nearest neighbour in binarized MNIST\@.}
        \label{fig:mnist_samples}
    \end{center}
\end{figure*}

We randomly sampled 100 digits from our best performing model from Table~\ref{tab:mnist_results} and compared them with their nearest neighbor in the training set (Figure~\ref{fig:mnist_samples}), to ensure that the generated samples are not simple memorization. Each row of digits uses a different mask that was seen at training time by the network.

\section{Conclusion}
We proposed MADE, a simple modification of autoencoders allowing them to be used as distribution estimators.
MADE demonstrates that it is possible to get direct, cheap estimates of high-dimensional joint probabilities, from a single pass through an autoencoder.
Like standard autoencoders, our extension is easy to vectorize and implement on GPUs.
MADE can evaluate high-dimensional probably distributions with better scaling than before, while maintaining state-of-the-art statistical performance.

\section*{Acknowledgments}
We thank Marc-Alexandre Côté for helping to implement NADE in Theano and the whole Theano~\cite{Bastien-Theano-2012,bergstra+al:2010-scipy} team of contributors. We also thank NSERC, Calcul Qu\'ebec and Compute Canada.

\bibliography{made}
\bibliographystyle{icml2015}

\appendix 

\twocolumn[
\icmltitle{MADE: Masked Autoencoder for Distribution Estimation.\\Supplementary Material}

\icmlauthor{Mathieu Germain}{mathieu.germain2@usherbrooke.ca}
\icmladdress{Université de Sherbrooke, Canada}
\icmlauthor{Karol Gregor}{karol.gregor@gmail.com}
\icmladdress{Google DeepMind}
\icmlauthor{Iain Murray}{i.murray@ed.ac.uk}
\icmladdress{University of Edinburgh, United Kingdom}
\icmlauthor{Hugo Larochelle}{hugo.larochelle@usherbrooke.ca}
\icmladdress{Université de Sherbrooke, Canada}

\icmlkeywords{deep learning, distribution estimation, autoregressive modelling, machine learning, autoencoder, unsupervised learning, generative model, ICML}

\vskip 0.3in
]

\newcommand{\tpm}[1]{\textit{\tiny{±#1}}}

\begin{table}
\caption{Negative log-likelihood and 95\% confidence intervals for Table~4 in the main document.}
\begin{center} 
\begin{tabular}{lccc}\toprule
& \multicolumn{2}{c}{MADE}                        & EoNADE \\
\cmidrule[.5pt]{2-3}
Dataset     & Fixed mask       &  Mask sampling   & 16 ord.\\
\midrule
Adult       &  13.12\tpm{0.05} &  13.13\tpm{0.05} &  13.19\tpm{0.04}\\
Connect4    &  11.90\tpm{0.01} &  11.90\tpm{0.01} &  12.58\tpm{0.01}\\
DNA         &  83.63\tpm{0.52} &  79.66\tpm{0.63} &  82.31\tpm{0.46}\\
Mushrooms   &   9.68\tpm{0.04} &   9.69\tpm{0.03} &   9.69\tpm{0.03}\\
NIPS-0-12   & 280.25\tpm{1.05} & 275.92\tpm{1.01} & 272.39\tpm{1.08}\\
Ocr-letters &  28.34\tpm{0.22} &  30.04\tpm{0.22} &  27.32\tpm{0.19}\\
RCV1        &  47.10\tpm{0.11} &  46.74\tpm{0.11} &  46.12\tpm{0.11}\\
Web         &  28.53\tpm{0.20} &  28.25\tpm{0.20} &  27.87\tpm{0.20}\\
\bottomrule
\label{tab:results_confidence_intervals}
\end{tabular}
\end{center} 
\end{table}

\begin{table}[!ht]
\begin{center}
\caption{Binarized MNIST negative log-likelihood and 95\% confidence intervals for Table~6 in the main document.}
\smallskip
\begin{tabular}{lc}\toprule
Model                     &  \\
\midrule
MADE 1hl (1 mask)     & 88.40\tpm{0.45} \\
MADE 2hl (1 mask)     & 89.59\tpm{0.46} \\
MADE 1hl (32 masks)   & 88.04\tpm{0.44} \\
MADE 2hl (32 masks)   & 86.64\tpm{0.44} \\
\bottomrule
\label{tab:mnist_results_confidence_intervals}
\end{tabular}
\end{center}
\end{table}

\end{document}